# Polish - English Speech Statistical Machine Translation Systems for the IWSLT 2014.

*Krzysztof Wołk, Krzysztof Marasek*

Multimedia Department
Polish Japanese Academy of Information Technology, Koszykowa 86, 02-008 Warsaw
kwolk@pja.edu.pl, kmarasek@pja.edu.pl

**Abstract**

This research explores effects of various training settings between Polish and English Statistical Machine Translation systems for spoken language. Various elements of the TED parallel text corpora for the IWSLT 2014 evaluation campaign were used as the basis for training of language models, and for development, tuning and testing of the translation system as well as Wikipedia based comparable corpora prepared by us. The BLEU, NIST, METEOR and TER metrics were used to evaluate the effects of data preparations on translation results. Our experiments included systems, which use lemma and morphological information on Polish words. We also conducted a deep analysis of provided Polish data as preparatory work for the automatic data correction and cleaning phase.

## 1. Introduction

Polish is one of the complex West-Slavic languages, which represents a serious challenge to any SMT system. The grammar of the Polish language, with its complicated rules and elements, together with a big vocabulary (due to complex declension) are the main reasons for its complexity (in Polish there are seven cases, three genders, animate and inanimate nouns, adjectives agreed with nouns in terms of gender, case and number and a lot of words borrowed from other languages which are often inflected similarly to those of Polish origin).

This greatly affects the data and data structure required for statistical models of translation. The lack of available and appropriate resources required for data input to SMT systems presents another problem. SMT systems should work best in specified, not too wide text domains and will not perform well for general use. Good quality parallel data, especially in a required domain has low availability. In general, Polish and English differ also in syntax. English is a positional language, which means that the syntactic order (the order of words in a sentence) plays a very important role, particularly due to limited inflection of words (e.g. lack of declension endings). Sometimes, the position of a word in a sentence is the only indicator of the sentence meaning. In the English sentence, the subject group comes before the predicate, so the sentence is ordered according to the Subject-Verb-Object (SVO) schema. In Polish, however, there is no specific word order imposed and the word order has no decisive influence on the understanding of the sentence. One can express the same thought in several ways, which is not possible in English. For example, the sentence „I just tasted a new orange juice." can be written in Polish as „Spróbowałem właśnie nowego soku pomarańczowego", or "Nowego soku pomarańczowego właśnie spróbowałem.", or "Właśnie spróbowałem nowego soku pomarańczowego.", or „Właśnie nowego soku pomarańczowego spróbowałem." Differences in potential sentence orders make the translation process more complex, especially when working on a phrase-model with no additional lexical information.

As a result starting point was much lower than for other languages, however our progress in last 3 years was faster than others [1,2]. The aim of this work is to create an SMT system for translation from Polish to English (and the other way round, i.e. from English to Polish) to address the IWSLT 2014 [3] evaluation campaign requirements. This paper is structured as follows: Section 2 explains the Polish data preparation. Section 3 presents the English language issues. Section 4 describes the translation evaluation methods. Section 5 presents the results. Lastly in Section 6 we summarize potential implications and ideas for future work.

## 2. Preparation of the Polish data

The Polish data in the TED talks (about 17 MB) include almost 2,5 million words that are not tokenized. The transcripts themselves are provided as pure text encoded with UTF-8 and the transcripts are prepared by the IWSLT team [4]. In addition, they are separated into sentences (one per line) and aligned in language pairs.

It should be emphasized that both automatic and manual preprocessing of this training information was required. The extraction of the transcription data from the provided XML files ensured an equal number of lines for English and Polish. However, some of the discrepancies in the text parallelism could not be avoided. These discrepancies are mainly repetitions of the Polish text not included in the English text.

Another problem was that TED 2013 data was full of errors. [5]. For the IWSLT 2014 we helped in repairing those errors in train, test and development sets. It was done semi-automatically by the usage of our tool described in [6]. We repaired spelling errors that artificially increased the dictionary size in Polish side of the corpora. Additionally we filtered out and repaired bi-sentences with odd nesting, such as:

Part A, Part A, Part B, Part B.

e.g.

*"Ale będę starał się udowodnić, że mimo złożoności, Ale będę starał się udowodnić, że mimo złożoności, istnieją pewne rzeczy pomagające w zrozumieniu. istnieją pewne rzeczy pomagające w zrozumieniu."*

Some parts (words or full phrases or even whole sentences) were duplicated. Furthermore, there were segments containing repetitions of whole sentences inside a single segment. For instance:

Sentence A. Sentence A.

e.g.

*"Zakumulują się u tych najbardziej pijanych i skąpych. Zakumulują się u tych najbardziej pijanych i skąpych."*

or

Part A, Part B, Part B, Part C
e.g.
*" Matka może się ponownie rozmnażać, ale jak wysoką cenę płaci, przez akumulację toksyn w swoim organizmie - przez akumulację toksyn w swoim organizmie - śmierć pierwszego młodego."*

Overall, in the train set we found about 7% of spelling errors and about 15% of insertion errors. Luckily such problems occur only on the Polish side of the corpora. In our opinion the pre-processing tools used to align the corpus were not adjusted for the Polish language. Cleaning those problems increases BLEU score by the factor of 1,5 – 2.

The number of unique Polish words and their forms was 144,115 and 59,296 English unique word forms. The disproportionate vocabulary sizes are also a challenge especially in translation from English to Polish.

Another problem is that the TED Talks do not have any specific domain. Statistical Machine Translation by definition works best when very specific domain data is used. The data we have is a mix of various, unrelated topics. This is most likely the reason why we cannot expect big improvements with this data and generally low scores in translation quality metrics.

There is not much focus on Polish in the campaign, so there is almost no additional data in Polish in comparison to a huge amount of data in, for example, French or German. At first we used perplexity measurement metrics to determine the data we obtained. Some of the data we were able to obtain from the OPUS [12] project page, some from another small projects and the rest was collected manually using web crawlers. We created those corpora and used them. What we created was:

- A Polish – English dictionary (bilingual parallel)
- Additional (newer) TED Talks data sets not included in the original train data (we crawled bilingual data and created a corpora from it) (bilingual parallel)
- E-books (monolingual PL + monolingual EN)
- Proceedings of UK House of Lords (monolingual EN)
- Subtitles for movies and TV series (monolingual PL)
- Parliament and senate proceedings (monolingual PL)
- Wikipedia Comparable Corpus (bilingual parallel)
- Euronews Comparable Corpus (bilingual parallel)
- Repository of PJIIT's diplomas (monolingual PL)
- Many PL monolingual data web crawled from main web portals like blogs, chip.pl, Focus newspaper archive, interia.pl, wp.pl, onet.pl, money.pl, Usenet, Termedia, Wordpress web pages, Wprost newspaper archive, Wyborcza newspaper archive, Newsweek newspaper archive, etc.

"Other" in the table below stands for many very small models merged together. In Table 1 we show the perplexity values of the obtained data with no smoothing (PPL in Table 1) as well as smoothed with the Kneser-Ney algorithm (PPL+KN in Table 1). We used the MITLM [29] toolkit for that evaluation. As an evaluation set we used dev2010 data, which was used for tuning. Its dictionary covers 2861 words.

EMEA are texts from the European Medicines Agency, KDE4 is a localization file of that GUI, ECB stands for European Central Bank corpus, OpenSubtitles [12] are movies and TV series subtitles, EUNEWS is a web crawl of the euronews.com web page and EUBOOKSHOP comes from bookshop.europa.eu. Lastly bilingual TEDDL is additional TED data. We ensured that this data was not overlapping with the test or development sets. As can be seen from the Table 1, all additional data has big perplexity values, so no astonishing improvements based only on data could be expected.

*Table 1*: Data Perplexities for dev2010 data set

| Data set | Dictionary | PPL | PPL + KN |
|---|---|---|---|
| Baseline train.en | 44,052 | 221 | 223 |
| EMEA | 30,204 | 1738 | 1848 |
| KDE4 | 34,442 | 890 | 919 |
| ECB | 17,121 | 837 | 889 |
| OpenSubtitles | 343,468 | 388 | 415 |
| EBOOKS | 528,712 | 405 | 417 |
| EUNEWS | 21,813 | 430 | 435 |
| NEWS COMM | 62,937 | 418 | 465 |
| EUBOOKSHOP | 167,811 | 921 | 950 |
| UN TEXTS | 175,007 | 681 | 714 |
| UK LORDS | 215,106 | 621 | 644 |
| NEWS 2010 | 279,039 | 356 | 377 |
| GIGAWORD | 287,096 | 582 | 610 |
| DICTIONARY | 39,214 | 8629 | 8824 |
| OTHER | 13,576 | 492 | 499 |
| WIKIPEDIA | 682,276 | 9131 | 9205 |
| NEWSPAPERS | 608,186 | 10066 | 10083 |
| WEB PORTALS | 510,240 | 731 | 746 |
| BLOGS | 76,697 | 3481 | 3524 |
| USENET | 733,619 | 8019 | 8034 |
| DIPLOMAS | 353,730 | 32345 | 32582 |
| TEDDL | 47,015 | 277 | 277 |

WIKIPEDIA and EUNEWS are parallel corpora extracted by us from comparable corpora. We were able to obtain 4,498 topic-aligned articles from the Euronews and about 1M from the Wikipedia. The Wikipedia corpus was about 104MB in size and contained 475,470 parallel sentences. Its first version was acknowledged as permissible data for the IWSLT 2014 evaluation campaign. The Euronews corpora contained 1,617 bi-sentences.

In order to extract the parallel sentence pairs we decided to facilitate Yalign Tool [26]. The Yalign tool was designed in order to automate parallel text mining process by finding sentences that are close translation matches from the comparable corpora. This opened up avenues for harvesting parallel corpora from sources like translated documents and the web. What is more Yalign is not limited to any language pair. But creation of own alignment models for two required languages is necessary.

The Yalign tool was implemented using a sentence similarity metric that produces a rough estimate (a number between 0 and 1) of how likely it is for two sentences to be a translation of each other. Additionally it uses a sequence aligner, that produces an alignment that maximizes the sum of the individual (per sentence pair) similarities between two documents. Yalign's main algorithm is actually a wrapper before standard sequence alignment algorithm [26].

For the sequence alignment Yalign uses a variation of the Needleman-Wunch algorithm [27] to find an optimal alignment between the sentences in two given documents. The algorithm has polynomial time worst-case complexity and it produces an optimal alignment. Unfortunately it can't handle alignments that cross each other or alignments from two sentences into a single one [27].

Since the sentence similarity is a computationally expensive operation, the implemented variation of the Needleman-Wunch algorithm uses A* approach to explore the search space instead of using the classical dynamic programming method that would require N * M calls to the sentence similarity matrix.

After the alignment, only sentences that have a high probability of being translations are included in the final alignment. The result is filtered in order to deliver high quality alignments. To do this, a threshold value is used, such that if the sentence similarity metric is low enough the pair is excluded.

For the sentence similarity metric the algorithm uses a statistical classifier's likelihood output and adapts it into the <0,1> range.

The classifier must be trained in order to determine if a pair of sentences is translation of each other or not. The particular classifier used in the Yalign project was a Support Vector Machine. Besides being excellent classifier, SVMs can provide a distance to the separation hyperplane during classification, and this distance can be easily modified using a Sigmoid Function to return likelihood between 0 and 1 [28].

The use of a classifier means that the quality of the alignment depends not only on the input but also on the quality of the trained classifier.

To train the classifier a good quality parallel data was necessary as well as a dictionary with translation probability included. For this purposes we used TED talks [3] corpora enhanced by us during the IWSLT'13 Evaluation Campaign [5]. In order to obtain a dictionary we trained a phrase table and extracted 1-grams from it. We used the MGIZA++ tool for word and phrase alignment. The lexical reordering was set to use the msd-bidirectional-fe method and the symmetrisation method was set to grow-diag-final-and for word alignment processing [5].

Before use of a training translation model, preprocessing that included removal of long sentences (set to 80 words) had to be performed. The Moses toolkit scripts [7] were used for this purpose.

The final processing corpus included 185,527 lines from the Polish to English corpus. However, the disproportionate vocabulary sizes remained. One of the solutions to this problem (according to work of Bojar [10]) was to use stems instead of surface forms in order to reduce the Polish vocabulary size. Such a solution also requires a creation of an SMT system from Polish stems to plain Polish. Subsequently, we used PSI-TOOLKIT [9] to convert each Polish word into a lemma. The toolkit is a tool chain for automatic processing of Polish language and to lesser extent other languages like English, German, French, Spanish and Russian (with the focus on machine translation). The tool chain includes segmentation, tokenization, lemmatization, shallow parsing, deep parsing, rule-based machine translation, statistical machine translation, automatic generation of inflected forms from lemma sequences and automatic post edition. The toolkit was used as an additional information source for the SMT system preparation. It can be also used as a first step for implementing a factored SMT system that, unlike a phrase-based system, includes morphological analysis, translation of lemmas and features as well as generation of surface forms. Incorporating additional linguistic information should effectively improve translation performance [8].

### 2.1. Polish lemma extraction

As previously mentioned, lemma extracted from Polish words are used instead of surface forms to overcome the problem of the huge difference in vocabulary sizes. For Polish lemma extraction, a tool chain that included tokenization and lemmatization from PSI-TOOLS was used.

These tools used in sequence provide a rich output that includes a lemma form of the tokens, prefixes, suffixes and morphosyntactic tags. Unfortunately unknown words like names or abbreviations or numbers, etc. are lost in the process. Also capitalization as well as punctuation does not remain. To preserve this relevant information we implemented a specialized tool that basing on differences between input and output of the PSI-TOOLS restored most of the lost information. The lemmatized version of the Polish training data was reduced to 36,065 unique words and the polish language model was also reduced from 156,970 to 32,873 unique words. The results of this work are presented in Table 2 and in Table 3. Each experiment was done only on the baseline data sets in PL->EN and EN->PL direction. The system settings are described in Chapter 5. The year column shows the test set that was used in the experiment, if a year has L suffix in means that it is lemmatized version of the baseline system.

*Table 2*: PL Lemma to EN translation results

| YEAR | BLEU | NIST | TER | MET |
| --- | --- | --- | --- | --- |
| 2010 | 16,70 | 5,70 | 67,83 | 49,31 |
| 2010L | 13,33 | 4,68 | 70,86 | 46,18 |
| 2011 | 20,40 | 5,71 | 62,99 | 53,13 |
| 2011L | 16,21 | 5,11 | 67,16 | 49,64 |
| 2012 | 17,22 | 5,37 | 65,96 | 49,72 |
| 2012L | 13,29 | 4,64 | 69,59 | 45,78 |
| 2013 | 18,16 | 5,44 | 65,50 | 50,73 |
| 2013L | 14,81 | 4,88 | 68,96 | 47,98 |
| 2014 | 14,71 | 4,93 | 68,20 | 47,20 |
| 2014L | 11,63 | 4,37 | 71,35 | 44,55 |

*Table 3*: EN to PL Lemma translation results

| YEAR | BLEU | NIST | TER | MET |
| --- | --- | --- | --- | --- |
| 2010 | 9,95 | 3,89 | 74,66 | 32,62 |
| 2010L | 12,98 | 4,86 | 68,06 | 40,19 |
| 2011 | 12,56 | 4,37 | 70,13 | 36,23 |
| 2011L | 16,36 | 5,40 | 62,96 | 44,86 |
| 2012 | 10,77 | 3,92 | 75,79 | 33,80 |
| 2012L | 14,13 | 4,83 | 69,76 | 41,52 |
| 2013 | 10,96 | 3,91 | 75,95 | 33,85 |
| 2013L | 15,21 | 5,02 | 68,17 | 42,58 |
| 2014 | 9,29 | 3,47 | 82,58 | 31,15 |
| 2014L | 12,35 | 4,44 | 75,27 | 39,12 |

Our experiments show that lemma translation to EN in each test set decreased the evaluation scores, contrary translation from EN to lemma for each set increased the translation quality. Such solution requires also training of a system from lemma into PL in order to restore proper surface forms of the words. We trained such system as well and evaluated it on official tests sets from years 2010-2014 and tuned on 2010 development data. The results for that system are presented in Table 4. Even that the scores are relatively high the results do not seem to be satisfactory enough to provide overall improvement of EN-LEMMA-PL pipeline over direct translation from EN to PL.

*Table 4*: Lemma to PL translation results

| YEAR | BLEU | NIST | TER | MET |
|---|---|---|---|---|
| 2010 | 41,14 | 8,72 | 31,28 | 65,25 |
| 2011 | 41,68 | 8,68 | 30,64 | 65,99 |
| 2012 | 38,87 | 8,38 | 32,23 | 64,18 |
| 2013 | 40,27 | 8,30 | 31,67 | 64,44 |
| 2014 | 37,78 | 8,01 | 33,17 | 62,78 |

To confirm our prediction we conducted additional experiment in which the English sentences were first translated into lemma and secondly we translated lemma into Polish surface forms. The results of such combined translation are showed in Table 5. They decrease the translation quality in comparison to direct translation from EN to PL. What is more by lemmatizing PL we lost much significant information. As a part of the future work we intend to lemmatize only not very common words, but we are still aware of that most of the Polish words will appear quire rare due to many word forms. We anticipate that most of the words will be replaced by lemmas. Unfortunately also the quality of lemma to surface is of low quality. The Polish declension is complex e.g. sometimes even a steam is changed doe to phonetic/phontactic rules.

*Table 5*: EN -> PL Lemma -> PL pipeline translation

| YEAR | BLEU | NIST | TER | MET |
|---|---|---|---|---|
| 2010 | 7,47 | 3,45 | 76,17 | 29,16 |
| 2011 | 9,67 | 3,84 | 72,45 | 32,25 |
| 2012 | 8,26 | 3,39 | 78,40 | 29,60 |
| 2013 | 8,83 | 3,54 | 77,11 | 30,61 |
| 2014 | 6,98 | 3,10 | 83,81 | 27,71 |

## 3. English Data Preparation

The preparation of the English data was definitively less complicated than for Polish. We developed a tool to clean the English data by removing foreign words, strange symbols, etc. Compare to Polish, the English data contained significantly less errors. Nonetheless some problems needed to be removed, most problematic were translations into languages other than English and strange UTF-8 symbols. We also found few duplications and insertions inside single segments.

## 4. Evaluation Methods

Metrics are necessary to measure the quality of translations produced by the SMT systems. For this, various automated metrics are available to compare SMT translations to high quality human translations. Since each human translator produces a translation with different word choices and orders, the best metrics measure SMT output against multiple reference human translations. For scoring purposes we used four well-known metrics that show high correlation with human judgments. Among the commonly used SMT metrics are: Bilingual Evaluation Understudy (BLEU), the U.S. National Institute of Standards & Technology (NIST) metric, the Metric for Evaluation of Translation with Explicit Ordering (METEOR), and Translation Error Rate (TER).

According to Koehn, BLEU [11] uses textual phrases of varying length to match SMT and reference translations. Scoring of this metric is determined by the weighted averages of those matches. [13]

To encourage infrequently used word translation, the NIST [13] metric scores the translation of such words higher and uses the arithmetic mean of the n-gram matches. Smaller differences in phrase length incur a smaller brevity penalty. This metric has shown advantages over the BLEU metric.

The METEOR [13] metric also changes the brevity penalty used by BLEU, uses the arithmetic mean like NIST, and considers matches in word order through examination of higher order n-grams. These changes increase score based on recall. It also considers best matches against multiple reference translations when evaluating the SMT output.

TER [14] compares the SMT and reference translations to determine the minimum number of edits a human would need to make for the translations to be equivalent in both fluency and semantics. The closest match to a reference translation is used in this metric. There are several types of edits considered: word deletion, word insertion, word order, word substitution, and phrase order.

## 5. Experimental Results

A number of experiments were performed to evaluate various versions for our SMT systems. The experiments involved a number of steps. Processing of the corpora was accomplished, including tokenization, cleaning, factorization, conversion to lower case, splitting, and a final cleaning after splitting. Training data was processed, and the language model was developed. Tuning was performed for each experiment. Lastly, the experiments were conducted.

The baseline system testing was done using the Moses open source SMT toolkit with its Experiment Management System (EMS) [15]. The SRI Language Modeling Toolkit (SRILM) [19] with an interpolated version of the Kneser-Key discounting (interpolate –unk –kndiscount) was used for 5-gram language model training. We used the MGIZA++ tool for word and phrase alignment. KenLM [17] was used to binarize the language model, with a lexical reordering set to use the msd-bidirectional-fe model. Reordering probabilities of phrases are conditioned on lexical values of a phrase. It considers three different orientation types on source and target phrases like monotone(M), swap(S) and discontinuous(D). The bidirectional reordering model adds probabilities of possible mutual positions of source counterparts to current and following phrases [18]. MGIZA++ is a multi-threaded version of the well-known GIZA++ tool [20]. The symmetrization method was set to grow-diag-final-and for word alignment processing. First two-way direction alignments obtained from GIZA++ were intersected, so only the alignment points that occurred in both alignments remained. In the second phase, additional alignment points existing in their union were added. The growing step adds potential alignment points of unaligned words and neighbors. Neighborhood can be set directly to left, right, top or bottom, as well as to diagonal (grow-diag). In the final step, alignment points between words from which at least one is unaligned are added (grow-diag-final). If the grow-diag-final-and method is used, an alignment point between two unaligned words appears. [15]

We conducted about a hundred of experiments using test and development 2010 data to determine the best possible translation settings from Polish to English and the reverse. For experiments we used Moses SMT with Experiment Management System (EMS) [15]. Starting from baseline (BLEU: 16,70) system tests, we raised our score through extending the language model with more data and by interpolating it linearly. We determined that not using lower casing, changing maximum sentence length to 95, maximum phrase length to 6 improves the BLEU score. Additionally we changed the language model order from 5 to 6 and changed the discounting method from Kneser-Ney to Witten-Bell. Those setting proved to increase translation quality for PL-EN language pair in [5]. In the training part, we changed the lexicalized reordering method from msd-bidirectional-fe to hier-mslr-bidirectional-fe. The system was also enriched with Operation Sequence Model (OSM) [21]. The motivation for OSM is that it provides phrase-based SMT models the ability to memorize dependencies and lexical triggers, it can search for any possible reordering, and it has a robust search mechanism. Additionally, OSM takes source and target context into account, and it does not have the spurious phrasal segmentation problem. The OSM is valuable especially for the strong reordering mechanism. It couples translation and reordering, handles both short and long distance reordering, and does not require a hard reordering limit [21]. What is more we used Compound Splitting feature [8]. Tuning was done using MERT tool with batch-mira feature and n-best list size was changed from 100 to 150. This setting and language models produced the score of BLEU equal to 21,57. Lastly we used all parallel data we were able to obtain. We adapted it using Modified Moore Levis Filtering [8]. From our experiments we conducted that best results are obtained when sampling about 150,000 bi-sentences from in-domain corpora and by using filtering after the word alignment. The ratio of data to be kept was set to 0,8 obtaining our best score equal to 23,74.

Because of a much bigger dictionary, the translation from EN to PL is significantly more complicated. Our baseline system score was 9,95 in BLEU. Similarly to PL-EN direction we determined that not using lower casing, changing maximum sentence length to 85, maximum phrase length to 7 improves the BLEU score. Additionally we set the language model order from 5 to 6 and changed the discounting method from Kneser-Ney to Witten-Bell. In the training part, we changed the lexicalized reordering method from msd-bidirectional-fe to tgttosrc. The system was also enriched with Operation Sequence Model (OSM). What is more we used Compund Splitting feature and we did punctuation normalization. Tuning was done using MERT tool with batch-mira feature and n-best list size was changed from 100 to 150. Training a hierarchical phrase-based translation model also improved results in this translation scenario [16].

This setting and language models produced the score of BLEU equal to 19,81. Lastly we used all parallel data we were able to obtain. We adapted it using Modified Moore Levis Filtering [8]. From our experiments we conducted that best results are obtained when sampling about 150,000 bi-sentences from in-domain corpora and by using filtering after the word alignment. The ratio of data to be kept was set to 0,9 obtaining our best score equal to 22,76.

*Table 6*: Polish-to-English translation

| System | Year | BLEU | NIST | TER | METEOR |
|---|---|---|---|---|---|
| BASE | 2010 | 16,70 | 5,70 | 67,83 | 49,31 |
| BEST | 2010 | 23,74 | 6,25 | 54,63 | 57,06 |
| BASE | 2011 | 20,40 | 5,71 | 62,99 | 53,13 |
| BEST | 2011 | 28,00 | 6,61 | 51,02 | 61,23 |
| BASE | 2012 | 17,22 | 5,37 | 65,96 | 49,72 |
| BEST | 2012 | 23,15 | 5,55 | 56,42 | 56,49 |
| BASE | 2013 | 18,16 | 5,44 | 65,50 | 50,73 |
| BEST | 2013 | 28,62 | 6,71 | 57,10 | 58,48 |
| BASE | 2014 | 14,71 | 4,93 | 68,20 | 47,20 |
| BEST | 2014 | 18,96 | 5,56 | 64,59 | 51,29 |

The experiments on our best systems were conducted with the use of the test data from years 2010-2014. These results are showed in Table 6 and Table 7, respectively, for the Polish-to-English and English-to-Polish translations. They are measured by the BLEU, NIST, TER and METEOR metrics. Note that a lower value of the TER metric is better, while the other metrics are better when their values are higher.

*Table 7*: English-to-Polish translation

| System | Year | BLEU | NIST | TER | METEOR |
|---|---|---|---|---|---|
| BASE | 2010 | 9,95 | 3,89 | 74,66 | 32,62 |
| BEST | 2010 | 22,76 | 5,83 | 60,23 | 49,18 |
| BASE | 2011 | 12,56 | 4,37 | 70,13 | 36,23 |
| BEST | 2011 | 29,20 | 6,54 | 55,02 | 51,48 |
| BASE | 2012 | 10,77 | 3,92 | 75,79 | 33,80 |
| BEST | 2012 | 26,33 | 5,93 | 60,88 | 47,85 |
| BASE | 2013 | 10,96 | 3,91 | 75,95 | 33,85 |
| BEST | 2013 | 26,61 | 5,99 | 59,94 | 48,44 |
| BASE | 2014 | 9,29 | 3,47 | 82,58 | 31,15 |
| BEST | 2014 | 16,59 | 4,48 | 73,66 | 38,85 |

## 6. Discussion & Conclusions

Several conclusions can be drawn from the experimental results presented here. Automatic and manual cleaning of the training files has some positive impact, among the variations of the experiments [5]. Obtaining and adapting additional bi-lingual and monolingual data produced the biggest influence on the translation quality itself. In each direction using OSM and adapting training and tuning parameters was necessary and it could not be simply replicated from other experiments.

What was uncommon and surprising the punctuation normalization and usage of the hierarchical phrase model improved the quality only in translation into the Polish language and had negative results in opposite direction experiments.

What is more, converting Polish surface forms of words to lemma reduces the Polish vocabulary, which should improve the English-to-Polish translation performance and opposite. The Polish to English translation typically outscores the English to Polish translation, even on the same data. It is also what we would expect in our experiments with lemma, nonetheless our initial assumptions were not confirmed in empirical tests.

Several potential opportunities for future work are of interest. Additional experiments using extended language models are warranted to determine if this improves SMT scores. We are also interested in developing some more web crawlers in order to obtain additional data that would most likely prove useful. What is more, the Wikipedia corpus we created is still very noisy. We are currently working on cleaning it semi-automatically.

In future we intend to try clustering the training data into word classes in order to obtain smoother distributions and better generalizations. Using class-based models was shown to be useful when translating into morphologically rich languages like Polish [23]. We are also planning on using Unsupervised Transliteration Models, that proved to be quite useful in MT for translating OOV words, for disambiguation and for translating closely related languages [24]. This feature would most likely help us overcome difference in the vocabulary size, especially when translating into PL. Using a Fill-up combination technique (instead of interpolation) that is useful when the relevance of the models is known a priori: typically, when one is trained on in-domain data and the others on out-of-domain data is also in our interests [25].

Neural machine translation is a recently proposed approach to machine translation. Unlike the traditional statistical machine translation, the neural machine translation aims at building a single neural network that can be jointly tuned to maximize the translation performance. The models proposed recently for neural machine translation often belong to a family of encoder-decoders and consists of an encoder that encodes a source sentence into a fixed-length vector from which a decoder generates a translation. We would like to test such methodology on PL-EN language pair in accordance to [22].

## 7. Acknowledgements

This work is supported by the European Community from the European Social Fund within the Interkadra project UDA-POKL-04.01.01-00-014/10-00 and Eu-Bridge 7th FR EU project (grant agreement n°287658).

## References

[1] M. Cetollo, J. Niehues, S. Stuker, L. Bentivogli, M. Federico, "Overview of the IWSLT2012 Evaluation Campaign", in Proceedings of the 10th International Workshop on Spoken Language Translation (IWSLT), Hong Kong, China, 2012

[2] M. Cetollo, J. Niehues, S. Stuker, L. Bentivogli, M. Federico, "Report on the 10$^{th}$ IWSLT Evaluation Campaign", in Proceedings of the 10th International Workshop on Spoken Language Translation (IWSLT), Heidelberg, Germany, 2013

[3] IWSLT2014 Evaluation Campaign, http://workshop2014.iwslt.org/

[4] https://sites.google.com/site/iwsltevaluation2014/mt-track

[5] K. Wołk, K. Marasek, "Polish – English Speech Statistical Machine Translation Systems for the IWSLT 2013", in Proceedings of the 10th International Workshop on Spoken Language Translation (IWSLT), Heidelberg, Germany, 2013

[6] K. Wołk, K. Marasek, „A Sentence Meaning Based Alignment Method for Parallel Text Corpora Preparation", Advances in Intelligent Systems and Computing volume 275, Madeira Island, Portugal, 2014, pp. 107-114

[7] P. Koehn, H. Hoang, A. Birch, C. Callison-Burch, M. Federico, N. Bertoldi, B. Cowan, W. Shen, C. Moran, R. Zens, C. Dyer, R. Bojar, A. Constantin, E. Herbst, "Moses: Open Source Toolkit for Statistical Machine Translation", in Proceedings of the ACL 2007 Demo and Poster Sessions, Prague, June 2007, pp. 177–180

[8] K. Wołk, K. Marasek, „Polish -English Statistical Machine Translation of Medical Texts", New Research in Multimedia and Internet Systems, September 2014, pp. 169-177

[9] F. Graliński, K. Jassem, M. Junczys-Dowmunt, „PSI-Toolkit: Natural language processing pipeline", Computational Linguistics - Applications, Heidelberg: Springer 2012, pp. 27-39

[10] O. Bojar, "Rich Morphology and What Can We Expect from Hybrid Approaches to MT", Invited talk at International Workshop on Using Linguistic Information for Hybrid Machine Translation(LIHMT-2011), 2011

[11] P. Koehn, "What is a Better Translation?", Reflections on Six Years of Running Evaluation Campaigns, 2011

[12] J. Tiedemann, "Parallel Data, Tools and Interfaces in OPUS", in Proceedings of the 8th International Conference on Language Resources and Evaluation (LREC 2012), 2012, pp. 2214-2218

[13] K. Wołk, K. Marasek, „Enhanced Bilingual Evaluation Understudy", Lecture Notes on Information Theory, volume 2 number 2, 2014, pp.191-197

[14] M. Snover, B. Dorr, R. Schwartz, L. Micciulla, J. Makhoul, "A Study of Translation Edit Rate with Targeted Human Annotation", Proceedings of 7th Conference of the Assoc. for Machine Translation in the Americas, Cambridge, August 2006.

[15] K. Wołk, K. Marasek, „Real-Time Statistical Speech Translation", Advances in Intelligent Systems and Computing volume 275, Madeira Island, Portugal, 2014, pp.107-114

[16] D. Chiang, "Hierarchical Phrase-Based Translation", Computational Linguistics Volume 33, Number 2, 2007

[17] K. Heafield, "KenLM: Faster and smaller language model queries", Proceedings of Sixth Workshop on Statistical Machine Translation, Association for Computational Linguistics, 2011

[18] M. Costa-Jussa, J. Fonollosa, "Using linear interpolation and weighted reordering hypotheses in the Moses system", Barcelona, Spain, 2010

[19] A. Stolcke, "SRILM – An Extensible Language Modeling Toolkit", *INTERSPEECH*, 2002.

[20] Q. Gao, S. Vogel, "Parallel Implementations of Word Alignment Tool", *Software Engineering, Testing, and Quality Assurance for Natural Language Processing*, June 2008, pp. 49-57

[21] N. Durrani, H. Schmid, A. Fraser, "A Joint Sequence Model with Integrated Reordering", *Proceedings of the 49th Annual Meeting of the Association for Computational Linguistics*, Portland, Oregon, June 19-24, 2011, pp. 1045–1054,

[22] D. Bahdanau, K. Cho, Y. Bengio, "Neural Machine Translation by Jointly Learning to Align and Translate", arXiv cs.CL 1409.0473, 2014

[23] N. Durrani, P. Koehn, H. Schmid, A. Fraser, "Investigating the Usefulness of Generalized Word Representations in SMT", *Proceedings of the 25th Annual Conference on Computational Linguistics (COLING)*, Dublin, Ireland, August, 2014

[24] N. Durrani, P. Koehn, H. Hoang, H. Sajjad, "Integrating an Unsupervised Transliteration Model into Statistical Machine Translation", *EACL2014*, Gothenburg, Sweden, 2014

[25] *A. Bisazza, N. Ruiz, M. Federico,* "Fill-up versus Interpolation Methods for Phrase-based SMT Adaptation*", In Proceedings of IWSLT 2011*, 2011, pp. 136-143

[26] G. Berrotarán, R. Carrascosa, A. Vine, Yalign documentation, http://yalign.readthedocs.org/en/latest/

[27] G. Musso, Sequence Alignment (Needleman-Wunsch, Smith-Waterman), http://www.cs.utoronto.ca/~brudno/bcb410/lec2notes.pdf

[28] Thorsten Joachims, "Text Categorization with Support Vector Machines: Learning with Many Relevant Features", *Lecture Notes in Computer Science Volume 1398*, 1998, pp 137-142, 2005

[29] B. Hsu, J. Glass, "Interative Language Model Estimation: Efficient Data Structure & Algorithms", *In Proceedings Interspeech*, 2008